\RequirePackage{pdfpages}
\documentclass[conference]{IEEEtran}
\IEEEoverridecommandlockouts
\usepackage{cite}
\usepackage{amsmath,amssymb,amsfonts}
\usepackage{algorithmic}
\usepackage{graphicx}
\usepackage{textcomp}
\usepackage{booktabs}
\usepackage{authblk}
\usepackage{makecell}
\usepackage{afterpage}
\usepackage{lipsum}
\usepackage{titling}
\usepackage[letterpaper,left=54pt,right=54pt,bottom=54pt,top=54pt]{geometry}
\usepackage{xcolor}
\usepackage{soul}
\usepackage[normalem]{ulem}
\usepackage{subcaption}
\def\BibTeX{{\rm B\kern-.05em{\sc i\kern-.025em b}\kern-.08em
    T\kern-.1667em\lower.7ex\hbox{E}\kern-.125emX}}
\usepackage{titling}

\title{\fontsize{24pt}{28pt}\selectfont Feature Matching-Based Gait Phase Prediction for Obstacle Crossing Control of Powered Transfemoral Prosthesis}
\author{Jiaxuan Zhang$^1$, Yuquan Leng$^1$, Yixuan Guo$^1$, Chenglong Fu$^1$%
\thanks{This work was supported in part by the Science, Technology, and Innovation Commission of Shenzhen Municipality under Grants JCYJ20230807093407016 and KCXFZ20230731093401004, Guangdong Basic and Applied Basic Research Foundation under Grant 2023A1515110779, and National Natural Science Foundation of China under Grant 62403230 and 52175272, and National Key R\&D Program of China under Grant 2024YFC3082800, National Undergraduate Training Program for Innovation and Entrepreneurship under Grant X202514325008. (Corresponding author: Yixuan Guo. E-mail: guoyx@sustech.edu.cn.)\\ \indent \indent $^1$Key Laboratory of Biomimetic Robotics and Intelligent Systems, Southern University of Science and Technology, Shenzhen 518055, China.}\\
}

\date{}

\begin{document}

\maketitle
\thispagestyle{empty}
\begin{abstract}

For amputees with powered transfemoral prosthetics, navigating obstacles or complex terrain remains challenging. This study addresses this issue by using an inertial sensor on the sound ankle to guide obstacle-crossing movements. A genetic algorithm computes the optimal neural network structure to predict the required angles of the thigh and knee joints. A gait progression prediction algorithm determines the actuation angle index for the prosthetic knee motor, ultimately defining the necessary thigh and knee angles and gait progression. Results show that when the standard deviation of Gaussian noise added to the thigh angle data is less than 1, the method can effectively eliminate noise interference, achieving 100\% accuracy in gait phase estimation under 150 Hz, with thigh angle prediction error being 8.71\% and knee angle prediction error being 6.78\%. These findings demonstrate the method's ability to accurately predict gait progression and joint angles, offering significant practical value for obstacle negotiation in powered transfemoral prosthetics.

\end{abstract}

\begin{IEEEkeywords}
Lower limb prosthetics, cross obstacles, phase phase prediction
\end{IEEEkeywords}

\section{Introduction}

Obstacle crossing is a particularly challenging maneuver for individuals with above-knee amputations. Conventional passive fixed-knee prostheses require substantial compensatory movements and result in asymmetrical limb motion, which over extended periods can lead to muscle fatigue and other musculoskeletal disorders \cite{hobara2011lower,lee2021effect,berry2006microprocessor}. Passive adjustable-knee prostheses can minimize asymmetrical movements during locomotion on level ground and inclines \cite{hobara2013lower,mendez2011powered}; however, in complex terrain, such as crossing obstacles and going up and down stairs, the lack of active prosthetic assistance still poses a risk of amputees falling \cite{berry2006microprocessor,eveld2022factors,hafner2007evaluation}, and their movements still lack symmetry \cite{roerdink2012evaluating}. Later, active-assisted knee-ankle prostheses have been developed to meet the diverse movement requirements of amputees across various scenarios. These prostheses utilize predictive algorithms to generate gait trajectories and impedance parameters, which largely prevent compensatory movements, enabling amputees to navigate complex terrain smoothly and stably with gait patterns similar to those of healthy individuals \cite{Wang2022MotionIntention,lv2022datamined}.Nonetheless, during obstacle crossing, the inherent randomness of human movement and the non-rhythmic nature of specific events significantly increase the risk of falls, thereby elevating the danger for amputees when encountering unexpected events and obstacles in daily life \cite{hafner2015physical,mendez2020powered}.

 To address the challenge of helping above-knee amputees cross obstacles with active thigh prostheses, Shihao Cheng et al. \cite{cheng2024automatic} proposed a solution involving the installation of ultrasonic distance sensors on the anterior side of the prosthetic lower leg. This method modifies the motion trajectory based on real-time obstacle distance data received from the ultrasonic sensors, thereby enabling obstacle avoidance. However, the gait progression estimation method used in the sub-phase of the gait phase employs linear piecewise interpolation, which introduces a certain degree of error in the predicted progression.
  Joel Mendez et al. \cite{mendez2020powered} introduced an approach that utilizes the swinging angle of the residual limb's thigh to integrate and select the prosthetic knee joint's motion trajectory. Specifically, a larger backward swing angle of the residual limb's thigh prior to crossing an obstacle results in a greater maximum knee flexion angle, facilitating assisted obstacle crossing. However, when traversing higher or longer obstacles, this method requires the prosthesis wearer to perform unnatural excessive backward swinging of the prosthetic side thigh, thereby increasing the knee swing amplitude and extending the swing duration of the prosthesis \cite{cheng2024automatic}. 
 Thus, enabling natural movement during obstacle crossing is the focus of this study.

\begin{figure*}[ht]
    \centering
    \includegraphics[width=0.9\textwidth]{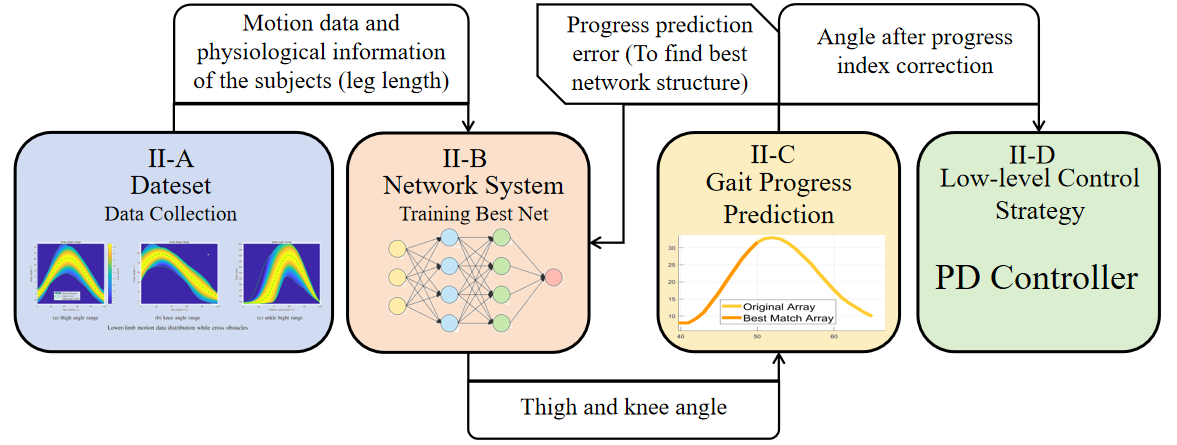}
    \caption*{ \small Fig. 1.\quad A simplified explanation of the implementation steps of the method. Section II-A explains the collection and processing of experimental data, Section II-B describes how to train and find the best neural network for predicting hip and knee joint angles also with gait phase estimation, Section II-C explains how to predict gait phase (gait progression), and finally Section II-D briefly describes how to convert the calculated angles into motor torques.}
    \label{fig:example}
\end{figure*}

In previous research conducted in our laboratory, a physical model of the prosthesis and a method for continuous-phase motion control have already been established\cite{Ma2022Apecie,yin2023environmental}. Therefore, in this paper, building upon the work of our predecessors, we focus solely on discussing trajectory prediction and phase estimation for the obstacle-crossing state. This paper does not delve into further discussion of other states and control methods.

This paper presents a neural network-based predictive control algorithm that assumes the prosthesis enters an obstacle-crossing state within the framework of a finite state machine. Utilizing vertical displacement data from an inertial measurement unit (IMU, MTw, Xsens, Netherlands) worn on the ankle of the sound leg, along with motion angle data from the previous gait phase of the residual limb on the prosthetic side, which is acquired using a motion capture system (Raptor-12, Motion Analysis, America), and the leg length of the subjects measured by hand. The neural network predicts the motion trajectory (angles) of the knee and hip joints of the prosthetic side. Subsequently, a dynamic time warping (DTW) based feature matching algorithm is employed to identify the progress of the gait phase. Based on previous laboratory research experience, this experiment utilizes only the angle of the thigh on the prosthetic side as the input variable for gait phase estimation \cite{chen2022piecewise,yin2023environmental}. Finally, the real-time computed phase progress index serves as the actuation index for the knee joint angle of the prosthesis, thereby enabling the execution of the knee joint angle and facilitating the realization of obstacle-crossing motion states of the prosthesis under data simulation.

\section{Method}
\begin{figure}[t]
\centering
    \begin{subfigure}{5.5cm}
    \centering
    \includegraphics[width=0.8\textwidth]{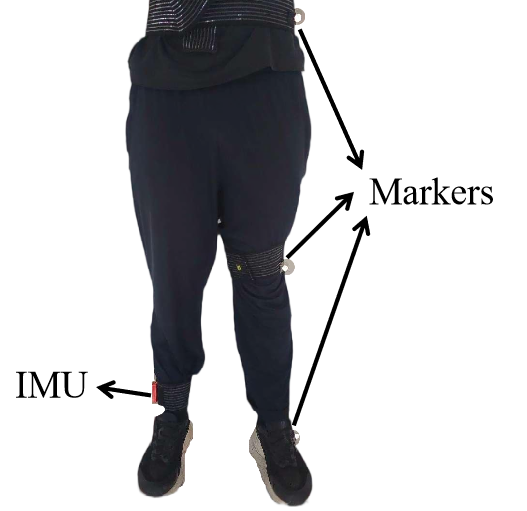}
    \caption{\small Sensor placement} \label{min2}
    \end{subfigure}
    \begin{subfigure}{9cm}
    \centering
    \includegraphics[width=0.8\textwidth]{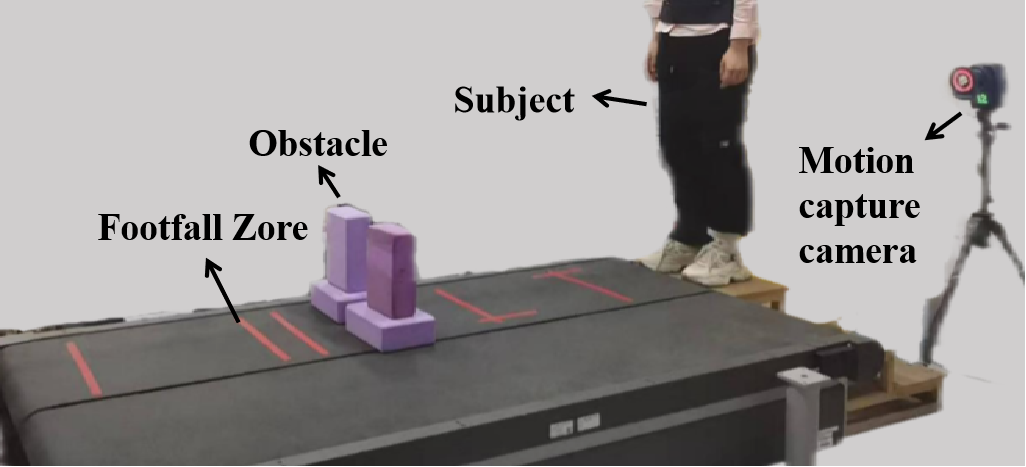}
    \caption{\small Data collection environment and locomotion protocol settings.} \label{min2}
    \end{subfigure}
    \caption*{\small Fig. 2.\quad Data collection settings. (a) An IMU is attached to the lateral side of the subject's right ankle. Motion capture markers are positioned at three key points: the most anterior point of the iliac crest, the most lateral aspect of the femoral epicondyle, and the outermost point of the ankle joint. (b) This Depicts the specific gait strategy planned for the subject, with red tape indicating the landing points for the subject's feet, and the obstacle composed of sponge.}
\end{figure}
The proposed control method is illustrated in Fig. 1. The rules and settings for obtaining these data are shown in Fig. 2. Our approach is predicated on the scenario where, during obstacle crossing, the amputee initially steps over with their sound leg. At this juncture, upon the sound leg's touchdown, the control system acquires the ankle joint's z-axis spatial coordinate data from the onset of the ascending phase to the touchdown phase of the leg that has already cleared the obstacle. Concurrently, it captures a segment of the prosthetic side's thigh motion trajectory prior to the sound leg's touchdown. This data is then fed into a neural network to generate the required hip and knee joint angles for the prosthetic side, serving as directives for the prosthetic limb's movement.

Subsequently, as the prosthesis begins to cross the obstacle (i.e., when the angle of the prosthetic thigh commences to increase from its previous minimum value and the residual thigh starts to swing forward), a gait progression algorithm is employed to calculate the gait progression within the current obstacle-crossing stride cycle based on the thigh's swing angle at the present moment. After completing the progression calculation, the motion of the prosthetic knee joint is actuated.
Section II-A describes the data collection procedures. The specific sensor configurations and data acquisition environments are depicted in Fig. 2, with the participant details presented in Table I. Section II-B introduces the neural network predictions and their applications, elucidating the process of obtaining the optimal network architecture using a genetic algorithm. Section II-C explains how to predict gait progression, and Section II-D briefly outlines the transformation of predicted angles into corresponding motion torques.
\subsection{Dataset}

In this study, kinematic data of healthy subjects crossing obstacles were gathered. Participants crossed obstacles first with their intact limb and then with their prosthetic limb. Obstacle heights varied from 150 mm to 300 mm, with foot-to-obstacle distances ranging from 100 mm to 300 mm. To promote natural obstacle crossing, only footfall zones were marked for guidance without strict foot placement regulation. Obstacle lengths were fixed between 140 mm and 270 mm. Participant details are in Table I, and sensor arrangements are shown in Fig. 2(a). Environmental conditions are depicted in Fig. 2(b).

\begin{table}[htbp]

\caption*{\makecell{TABLE I \\  Subject information}}
\begin{center}
\begin{tabular}{cccccc}
\toprule[1pt]
\vspace{1\baselineskip}
\makecell{\textbf{ID}} & \makecell{\textbf{Gender}} & \makecell{\textbf{Height } \\ \textbf{(cm)}} & \makecell{\textbf{Mass (kg)}} & \makecell{\textbf{Leg Length} \\ \textbf{ (cm)}} & \makecell{\textbf{Age (yrs)}} \\
\midrule
A & Male & 175 & 72 & 89 & 21 \\

B & Female & 170 & 65 & 85 & 20 \\

\bottomrule[1pt]
\end{tabular}
\label{tab:participant_info}
\end{center}
\end{table}

\begin{figure}[ht]
    \centering
    \includegraphics[width=0.45\textwidth]{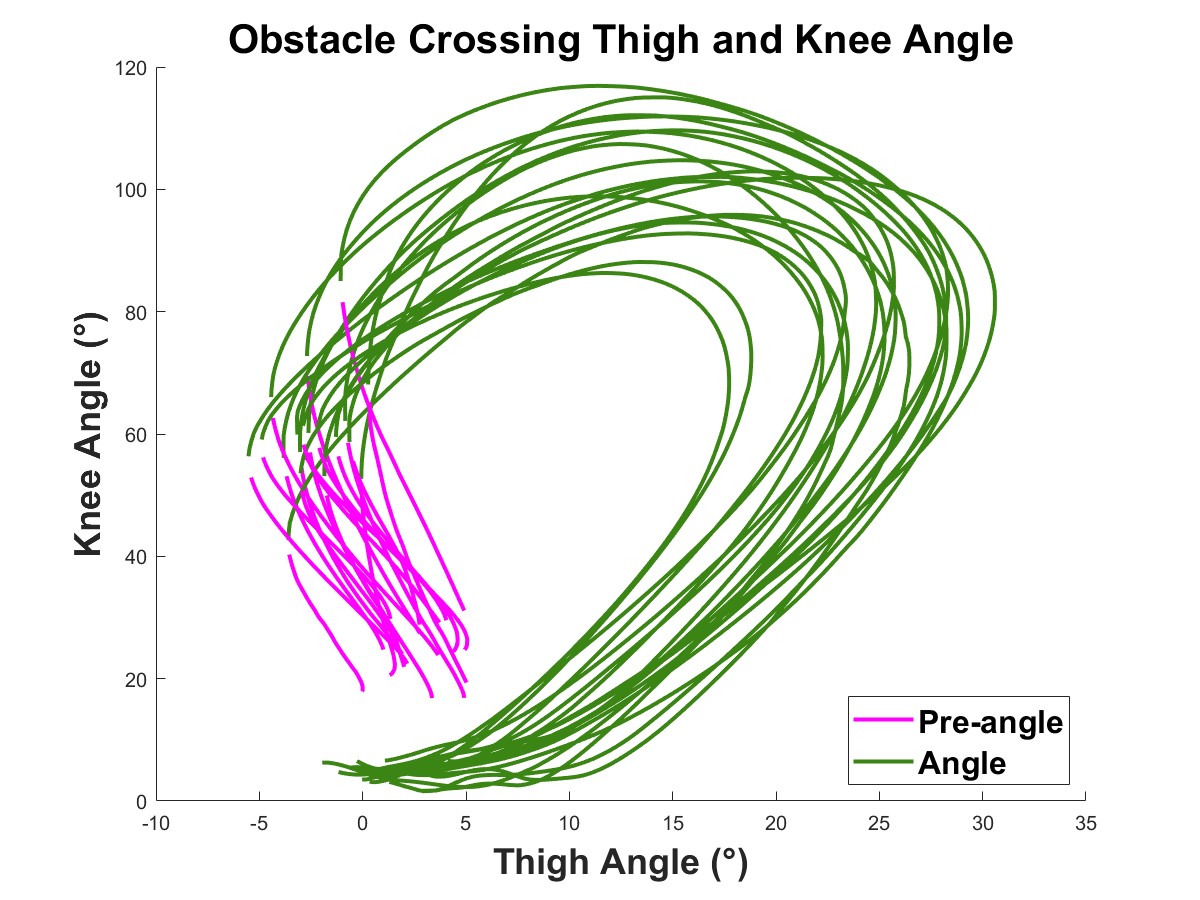}
    \caption*{\small Fig. 3.\quad Hip and knee angle correspondence during obstacle crossing. The magenta-colored angle data represents the data from the phase prior to the obstacle-crossing stage, while the green-colored angle data corresponds to the phase during the obstacle-crossing stage, which is the target data for prediction.}
    \label{fig:example}
\end{figure}

After the data collection is completed, the coordinate points are transformed and calculated into joint angles. The mapping relationship between the thigh and knee joint angles of the trailing leg (the leg that crosses the obstacle later, assumed to be the side with amputation) during obstacle crossing is shown in Fig. 3. This experiment divides the expected angle data (Angle) based on the minimum position of the thigh angle in this cycle to the subsequent minimum knee joint angle. Based on 20\% of the length of this data segment, the pre-obstacle crossing angle data (Pre-angle) is delineated from before the minimum thigh angle point. This pre-angle data serves as a corrective guide for subsequent obstacle crossing data prediction and as one of the inputs for the network. The ankle joint height data of the healthy side is cropped from the previous minimum point after entering the obstacle crossing cycle to the last minimum point.

\subsection{Network System}

Neural networks have shown promise in controlling prosthetic limbs, with several studies demonstrating their effectiveness. Kibria et al. \cite{kibria2024neurodynamic} utilized actor-critic networks to reduce gait asymmetry, while Kim et al. \cite{kim2022deeplearning} used deep learning to map gait trajectories for controlling a bionic leg. Mahmud et al. \cite{mahmud2021safe} explored neural-dynamics optimization for real-time joint trajectory generation, enhancing prosthetic adaptability.
T. Kevin Best et al. \cite{best2023data} proposed a kinematic control method for powered knee-ankle prostheses, achieving natural movement during the swing phase by combining impedance and kinematic control strategies. This hybrid approach illustrates the practicality and sophistication of using neural networks for prosthetic control during the swing phase.

This study designed a neural network to predict the angles of the thigh and knee joint  of a prosthetic leg during the obstacle-crossing phase. The network uses ankle joint height  from the healthy leg and predicted thigh angles from the prosthetic leg as inputs. It predicts the joint angles during the obstacle-crossing state. The system completes the predictions before the prosthetic leg lift-off the graund, avoiding delays. The basic logic of finding the best network is shown in Fig. 4.

\begin{figure}[htbp]
    \centering
    \includegraphics[width=0.95\linewidth]{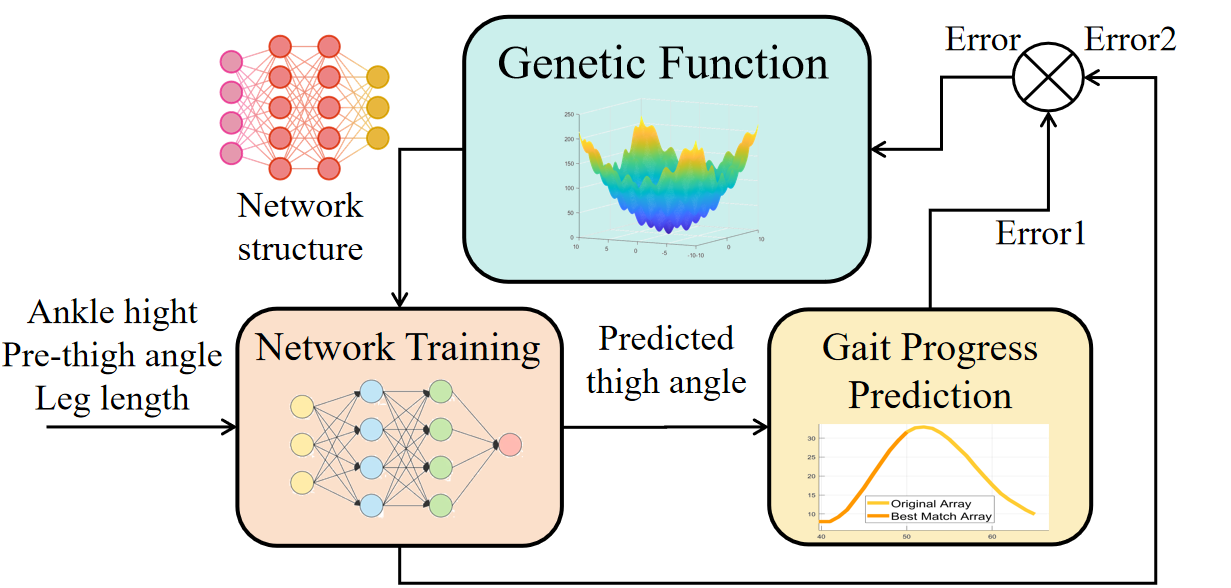}
    \caption*{\small Fig. 4. \quad Optimal Network Search Block Diagram. This illustrates obtaining the optimal network structure via a genetic function to optimize both gait cycle progress and joint angle trajectory predictions. "Angle height" denotes the height data of the healthy leg's ankle during obstacle crossing, while "Pre-thigh angle" refers to the angle data of the prosthetic leg before crossing. "Network structure" is the architecture passed from the genetic function to the network training function. "Thigh angle" represents the predicted thigh angle generated by the network. "Error1," "Error2," and "Error" correspond to the prediction errors of gait cycle progress, hip and knee joint angles, and their weighted sum, respectively.}
    \label{fig:enter-label}
\end{figure}

For training, processed data including ankle height and thigh angles were used. We employed attention mechanisms to modulate the focus of the output. Approximately 35\% of the data was randomly selected for validation. Training halted after 60 epochs without improvement in validation error, returning the network with the lowest loss.
Post-training, the predicted knee angles were locally adjusted to match the swing phase. This completed the construction and application of the neural network.

The study also explored the optimization network architecture inspired by Evolutionary Neural Networks (ENNs) \cite{salimans2017evolution}. A genetic function identified the optimal network structure, including layer types, counts, activation functions, dropout rates, and neuron numbers. The loss function of evolutionary algorithm combined RMSE of predictions and gait progression with an additional time-based component, which is armed to minimize computing time. The genetic algorithm had the following settings: population size of 20, three elites, 20 generations, crossover ratio of 0.8, uniform mutation function, mutation rate of 0.1.

\subsection{Phase Progress Estimation}

Accurate prediction of gait phases is vital for enhancing the functionality and user experience of lower-limb prostheses. Traditional time-based estimation (TBE) methods often struggle with dynamic changes in walking speed and terrain, resulting in inaccurate gait phase estimation \cite{kang2021real}. Recent studies have shown that machine learning algorithms can improve gait phase prediction in prosthetic legs by adapting to different walking conditions. For example, Best et al.\cite{best2023data} demonstrated a data-driven control strategy that effectively predicts gait phases and enhances natural movement for amputees.

In this paper, the phase estimation is based on feature matching from the data. Specifically, the relationship between the thigh angle of the prosthetic leg and time is recorded once the prosthetic thigh angle begins to increase from its minimum point, indicating the entry into the obstacle-crossing state. The thigh angle-time curve is then resampled to a fixed length. These resampled values are compared with the predicted thigh angle features, and the optimal match is determined by calculating the Dynamic Time Warping (DTW) distance \cite{dtw2020statistical,dtw2022series} between the two sequences. The index of the last matching segment is then returned and mapped to the corresponding knee joint angle index. This completes the mapping from the prosthetic thigh movement to the prosthetic knee joint movement. Below is a detailed explanation of the calculation of this mapping:

\subsubsection{Sequence Definition and Extension}

Given two sequences:
\begin{equation}
\mathbf{RS} = [RS_1, RS_2, \dots, RS_m], \quad 
\mathbf{CS} = [CS_1, CS_2, \dots, CS_n]
\end{equation}
\(RS\)  is the reference thigh angle trajectory curve calculated by the neural network, and \(CS\) is the true thigh angle trajectory of amputees after entering the obstacle crossing gait cycle. To improve alignment and reduce the impact of edge effects, both sequences are extended at the beginning with an additional transition sequence based on their characteristics.
Initially, the first three matching data segments will default to using a short extension. Subsequently, the optimal matching sequence determined from this calculation is used to extend the data selection within the Recurrence Sequence (RS) based on its position.
\begin{itemize}
    \item \text{Case 1: Short Extension (Sinusoidal Transition)} 
        If the number of computation iterations is less than three and the computed progress is within the first 30\% and exceeds 30\% but does not exceed five consecutive computations,, a sinusoidal transition with exponential decay is applied. The transition function is given by:
        \begin{equation}
        T_{\text{sin}}(t) = b + A \cdot \sin(2\pi t) \cdot e^{-k t}, \quad t \in [0, \text{num\_cycles}]
        \end{equation}
        where
         \( b \) is the baseline value, initialized as the first element of the sequence (\( RS_1 \) or \( CS_1 \)),
         \( A \) is the amplitude of oscillation 
         \( k \) is the decay factor controlling the rate of exponential damping,
         \( \text{num\_cycles} \) is the number of oscillation cycles.
     \item \text{Case 2: Long Extension (Linear Transition)} 
         If the returned progress is greater than 30\% and for five consecutive iterations., a linear monotonic transition is applied. The transition function adjusts the initial values based on the trend of the first few points and is given by:
        \begin{equation}
        T_{\text{lin}}(x) = b - \bigl(\text{slop} \cdot (x - 1)\bigr) \cdot e^{-c x}
        \end{equation}
        where
         \( b \) is the baseline value, initialized as the first element of the sequence (\( RS_1 \) or \( CS_1 \)),
         \( \text{slop} \) is the average slope, calculated from the first few points of the sequence:
        \[
        \text{slop} = \frac{(RS_2 - RS_1) + (RS_4 - RS_3)}{2} \quad \text{or}
        \]
        \[
         \quad \frac{(CS_2 - CS_1) + (CS_4 - CS_3)}{2}
        \]
        where \( c \) is the decay factor controlling the rate of damping.
    \item \text{Extended Sequences}
        Using the transitions defined above, the extended sequences are:
        \begin{equation}
        \begin{split}
           & \mathbf{RS'} = [T_{\text{sin}} \; \text{or} \; T_{\text{lin}}, \mathbf{RS}], \\
        &\mathbf{CS'} = [T_{\text{sin}} \; \text{or} \; T_{\text{lin}}, \mathbf{CS}]
        \end{split}
        \end{equation}
        The choice of \( T_{\text{sin}} \) or \( T_{\text{lin}} \) depends on the required extension length \( l_1 \), as described above.
\end{itemize}

\subsubsection{Optimized DTW Distance Calculation}

The sliding window extracts a sub-sequence of \( \mathbf{RS'} \) at position \( i \):
\begin{equation}
\mathbf{RS''}_{i} = \mathbf{RS'}_{1:i+n'+l_1-1} = [RS'_1, RS'_{2}, \dots, RS'_{i+n'+l_1-1}]
\end{equation}
To compute the DTW distance, the reference sequence \( \mathbf{RS''}_{i} \) is transformed using two vertical parameters (\( t_y, s_y \)) and a horizontal scaling factor (\( s_x \)), which modifies the length of the sliding window. The transformation function \( f(\mathbf{RS''}_{i}, \text{params}) \) is defined as:
\begin{equation}
f(\mathbf{RS''}_{i}, \text{params}) = s_y \cdot (\mathbf{RS''}_{i} - t_y)
\end{equation}
The horizontal scaling factor \( s_x \) adjusts the length of the sliding window:
\begin{equation}
\begin{split}
    & \quad \quad      \text{Scaled Window Length} = n' = s_x \cdot n, \\
     &    \text{with } s_x > 0 \text{ and } i + s_x \cdot n - 1 \leq m + l_1 \text{ and } n'>l_1
\end{split}
\end{equation}
where
 \( t_y \) is the vertical translation parameter, shifting the sequence's values along the \( Y \)-axis,
 \( s_y \) is the vertical scaling parameter, amplifying or compressing the sequence's values,
 \( s_x \) is the horizontal scaling parameter, changing the window's length,
 \( i \) is the starting position of the sliding window in \( \mathbf{RS''} \),
 \( n \) is the original length of the comparison sequence \( \mathbf{CS'} \),
 \( n' \) is the length of the scaled window \( \mathbf{RS''} \),
 \( m + l_1 \) is the total length of the extended reference sequence \( \mathbf{RS'} \).

The DTW distance is computed as:
\begin{equation}
\begin{split}
    & \quad \quad \text{DTW}(f(\mathbf{RS''}_{i}, \text{params}), \mathbf{CS'}) = \\
&\min \Biggl( \sum_{(i', j') \in P} \Bigl| f(RS''_{i'}, \text{params}) - CS'_{j'} \Bigr| \Biggr)
\end{split}
\end{equation}
where \( P \) is the warping path that aligns \( f(\mathbf{RS''}_{i}, \text{params}) \) and \( \mathbf{CS'} \), subject to the monotonicity and continuity constraints.

\subsubsection{Overall Optimization}

The overall optimization involves finding the window position \( i \) and the best transformation parameters that minimize the DTW distance:
\begin{equation}
\hat{i}, \hat{\text{params}} = \arg\min_{i=1, \dots, m-n+1} \; \text{DTW}(f(\mathbf{RS''}_{i}, \text{params}), \mathbf{CS'})
\end{equation}
where
 \( \hat{i} \) is the optimal starting position of the sliding window in \( \mathbf{RS'} \),
 \( \hat{\text{params}} = (\hat{t}_x, \hat{t}_y, \hat{s}_x, \hat{s}_y) \) are the optimal transformation parameters for \( \mathbf{RS''} \),
 \( \text{DTW}(\cdot, \cdot) \) is the DTW distance function defined over the transformed \( \mathbf{RS''} \) and \( \mathbf{CS'} \),
 \( f(\mathbf{RS''}_{i}, \text{params}) \) is the transformation function applied to \( \mathbf{RS''}_{i} \) with parameters \( \text{params} \).

Then we employ a heuristic algorithm to compute the optimal parameters.

\subsubsection{Best Match Index Calculation}

 Once the optimal window position \( \hat{i} \) is determined, the corresponding best-matching sequence in \( \mathbf{RS'} \) ends at \( \hat{i} + n' - 1 \). To convert this index into a progress value (\( \text{Progress} \)), the index must be normalized relative to the total length of the reference sequence \( \mathbf{RS} \).

\subsection{Low-Level Control Strategy}

Once the neural network has computed the anticipated knee joint angle, and the corresponding progress index has been determined through the gait cycle progression estimation algorithm, the desired angle can then be translated into the desired motor torque using common PD control methods. This part has been previously discussed in our earlier research\cite{yin2023environmental} and will not be elaborated further here.

\section{System Evaluation}

This paper employs a self-collected dataset for testing. A neural network is utilized to predict the joint angles of the prosthetic limb, followed by the application of the Pearson correlation coefficient ($r^2$) and Root Mean Squared Error (RMSE) to assess accuracy when compared with actual data. Offline tests are conducted to evaluate the phase estimation algorithm by simulating real-time updates of thigh angle data. IMU data and joint data are sent at a frequency of 25 to 150 Hz with added noise, and the phase progression algorithm receives data based on its own computational latency and the status of the received data as a reference for the calculation interval. This approach identifies matching positions within the predicted sequence to assess the accuracy of gait phase prediction, and compares the converted knee angles with actual data for evaluation.

\section{Discussion and Conclusion}

This paper proposes a control method for powered knee-ankle prostheses designed to assist amputees in crossing obstacles. Specifically, the method utilizes the thigh angle of the prosthetic side and the motion trajectory of the healthy-side ankle joint as inputs to a neural network, which predicts the motion trajectories of the hip and knee joints. Additionally, the thigh angle is used as an input to predict gait cycle progression in the obstacle-crossing control algorithm. The final trajectories of the thigh and knee joint angles are shown in Fig. 5. The thigh angle prediction accuracy is relatively low, but it does not affect the precision of gait phase prediction.

\begin{figure}[h]
    \centering
    \includegraphics[width=0.475\textwidth]{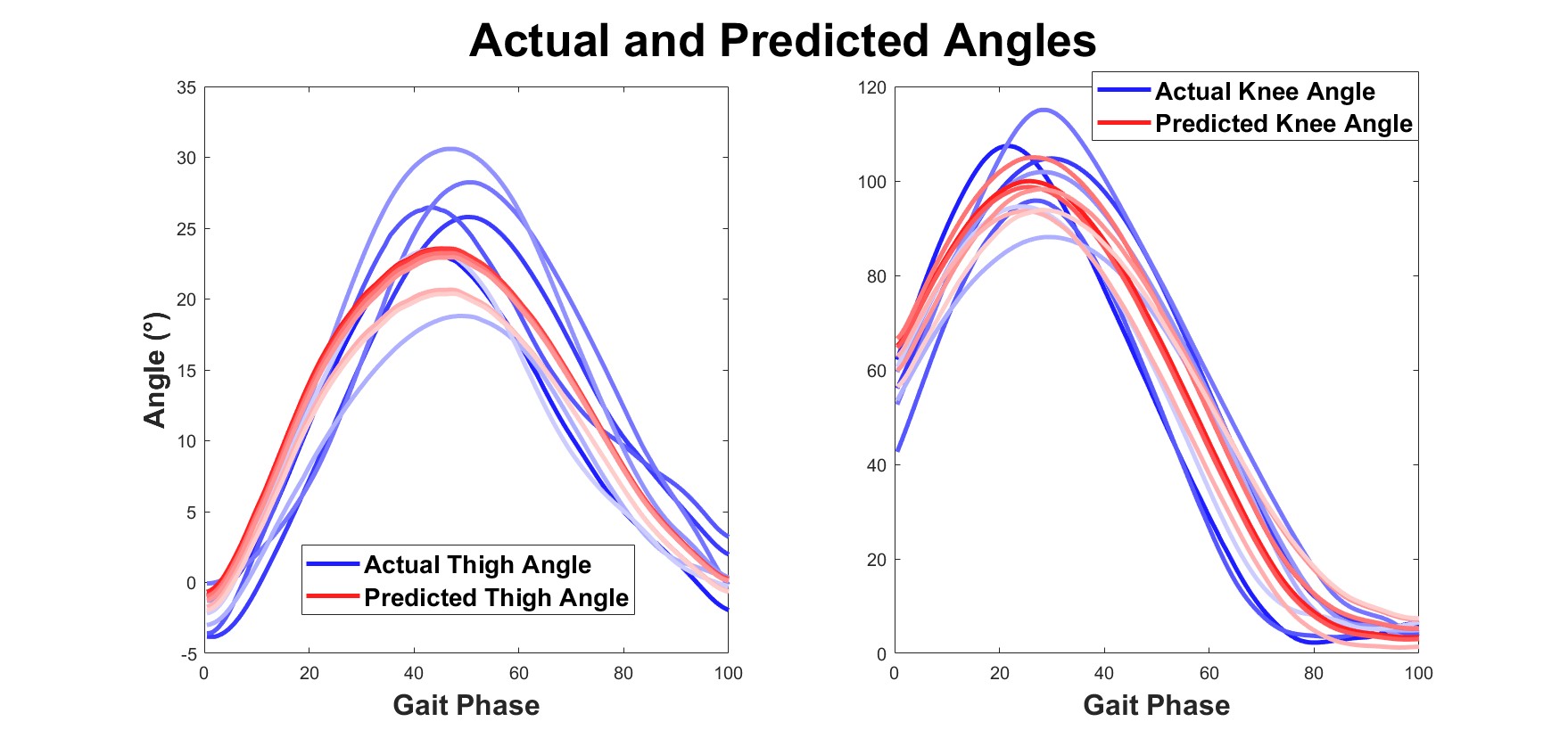}
    \caption*{\small Fig. 5. \quad Correspondence diagram of predicted angle and actual angle. The two colors from light to dark represent the correspondence of each group of real data and predicted data.}
    \label{fig:example}
\end{figure}

The overall accuracy of the proposed method was evaluated using a test dataset, where the root mean squared error (RMSE) for knee joint angles was calculated to be 6.78\%, and the RMSE error for thigh angles was 8.71\%. The Pearson correlation coefficients for the predicted knee and thigh angles were 0.985 and 0.963, respectively; for gait phase (gait progression) prediction,
tests indicate good sampling rate downward compatibility, with all sampling rates of gait phase progression accuracy being 100\% in tests ranging from 25 to 150 Hz. These simulation results demonstrate the high precision of the proposed method in predicting joint angles and gait phase, thereby validating its feasibility.

To assess the robustness and interference resistance of the control method in terms of gait phase progression and trajectory prediction, we applied post-processing Gaussian noise to the real-time thigh angle data input to the gait phase progression algorithm to simulate the impact of real outdoor environments on amputees and sensors. The standard deviation of the generated noise starts from 0.05, increases by 0.025, and has an upper limit of 3. Ultimately, by varying the noise production standard deviation, we obtained the images shown in Fig. 6, which demonstrate that the proposed method possesses a certain level of interference resistance. In the final test results, it can be observed that when the noise standard deviation is less than 1, the proposed method can effectively eliminate the interference of noise on gait cycle prediction. When the noise standard deviation is greater than 1 but less than 3, the method can also keep the RMSE of the gait phase progression within 2.4\%. Demonstrated robustness potential to real-world collisions.

Such simulation results have demonstrated that the method possesses a high level of robustness and adaptability.
\begin{figure}[h]
    \centering
    \includegraphics[width=0.475\textwidth]{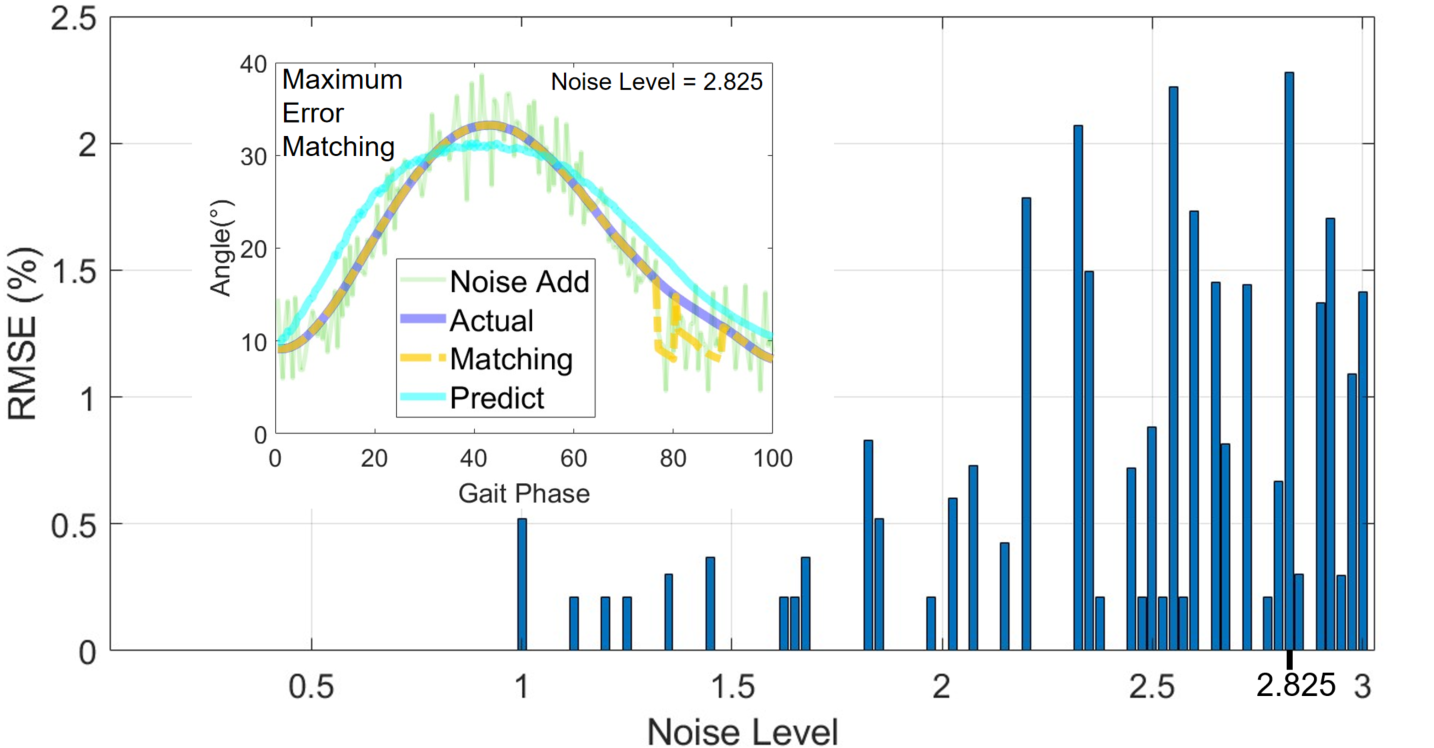}
    \caption*{\small Fig. 6. \quad Effects of noise on gait phase prediction. The blue column in the figure shows the prediction error RMSE as the noise level(gaussian noise standard deviation) increases. The small figure shows the matching details of the progress matching with the largest error in the data set when the noise level is 2.825. The lines in the legend are from top to bottom: real data with noise (CS), real data, matching data, and predicted data (RS).}
    \label{fig:example}
\end{figure}

This study's limitation is that tests and simulations were conducted on a laptop without real subject validation. The focus was on trajectory prediction and gait phase prediction during a single obstacle-crossing phase, potentially lacking adaptability for continuous gait phases. Future work will transfer the algorithm to real prostheses and test with amputees outdoors to assess control method feasibility and conduct comparative analyses with existing solutions. 

\section{Acknowledgement}
The authors would like to express their sincere appreciation to Tianyi Yu and Haizhou Yin for their assistance as subjects.

\bibliographystyle{IEEEtran}
\bibliography{referrences}

\end{document}